\documentclass[a4paper, 11pt]{article}
\usepackage[round]{natbib}
\newcommand{\fullname}{Adam D.\ Bull}
\newcommand{\thetitle}{Convergence rates of efficient global optimization 
algorithms}

\newcommand{\mscone}{90C26}
\newcommand{\msctwo}{68Q32}
\newcommand{\mscthree}{62C10}
\newcommand{\mscfour}{62L05}
\newcommand{\themsclass}{\mscone\ (Primary); \msctwo, \mscthree, \mscfour, \ 
(Secondary)}

\newcommand{\kwdone}{convergence rates}
\newcommand{\kwdtwo}{efficient global optimization}
\newcommand{\kwdthree}{expected improvement}
\newcommand{\kwdfour}{continuum-armed bandit}
\newcommand{\kwdfive}{Bayesian optimization}
\newcommand{\thekeywords}{\kwdone, \kwdtwo, \kwdthree, \kwdfour, \kwdfive}

\newcommand{\addressone}{Statistical Laboratory}
\newcommand{\addresstwo}{University of Cambridge}

\newcommand{\theemail}{a.bull@statslab.cam.ac.uk}

\newcommand{\theabstract}{
  In the efficient global optimization problem, we minimize an unknown 
  function $f$, using as few observations $f(x)$ as possible. It can be 
  considered a continuum-armed-bandit problem, with noiseless data, and simple 
  regret.  Expected-improvement algorithms are perhaps the most popular 
  methods for solving the problem; in this paper, we provide theoretical 
  results on their asymptotic behaviour.
  
  Implementing these algorithms requires a choice of Gaussian-process prior, 
  which determines an associated space of functions, its reproducing-kernel 
  Hilbert space (RKHS).  When the prior is fixed, expected improvement is 
  known to converge on the minimum of any function in its RKHS.  We provide 
  convergence rates for this procedure, optimal for functions of low 
  smoothness, and describe a modified algorithm attaining optimal rates for 
  smoother functions.

  In practice, however, priors are typically estimated sequentially from the 
  data.  For standard estimators, we show this procedure may never find the 
  minimum of $f$.  We then propose alternative estimators, chosen to minimize 
  the constants in the rate of convergence, and show these estimators retain 
  the convergence rates of a fixed prior.
}

\usepackage{amsmath, amsthm, amsfonts, mathtools, thmtools}
\usepackage{natbib, booktabs, tikz}
\usepackage[bookmarks, breaklinks, colorlinks, linkcolor=blue, citecolor=blue,
pdftitle={\thetitle}, pdfauthor={\fullname}]{hyperref}

\DeclarePairedDelimiter{\abs}{\lvert}{\rvert}
\DeclarePairedDelimiter{\norm}{\lVert}{\rVert}
\DeclareMathOperator{\spn}{span}
\newcommand{\N}{\mathbb{N}}
\newcommand{\C}{\mathbb{C}}
\newcommand{\R}{\mathbb{R}}
\renewcommand{\P}{\mathbb{P}}
\newcommand{\E}{\mathbb{E}}

\newcommand{\Cov}{\mathbb{C}\mathrm{ov}}

\declaretheorem{theorem}
\declaretheorem{lemma}
\declaretheorem{corollary}
\declaretheorem{definition}
\declaretheorem{assumption}

\newcommand{\acks}[1]{\subsection*{Acknowledgements} #1}
\begin{document}

\title{\thetitle
\footnotetext{\emph{Mathematics subject classification 2010.} \themsclass}
}
\author{\fullname\\\footnotesize \addressone\\\footnotesize \addresstwo\\
\footnotesize \theemail}
\date{}

\maketitle

\begin{abstract}
  \theabstract
\end{abstract}

\section{Introduction}
\label{sec:introduction}

Suppose we wish to minimize a continuous function $f:X \to \R$, where $X$ is a 
compact subset of $\R^d$. Observing $f(x)$ is costly (it may require a lengthy 
computer simulation or physical experiment), so we wish to use as few 
observations as possible. We know little about the shape of $f$; in particular 
we will be unable to make assumptions of convexity or unimodality.  We 
therefore need a {\em global} optimization algorithm, one which attempts to 
find a global minimum.

Many standard global optimization algorithms exist, including genetic 
algorithms, multistart, and simulated annealing 
\citep{pardalos_handbook_2002}, but these algorithms are designed for 
functions that are cheap to evaluate.  When $f$ is expensive, we need an {\em 
efficient} algorithm, one which will choose its observations to maximize the 
information gained.

We can consider this a continuum-armed-bandit problem \citep[and references 
therein]{srinivas_gaussian_2010}, with noiseless data, and loss measured by 
the simple regret \citep{bubeck_pure_2009}.  At time $n$, we choose a design 
point $x_n \in X$, make an observation $z_n = f(x_n)$, and then report a point 
$x_n^*$ where we believe $f(x_n^*)$ will be low.  Our goal is to find a 
strategy for choosing the $x_n$ and $x_n^*$, in terms of previous 
observations, so as to minimize $f(x_n^*)$.

We would like to find a strategy which can guarantee convergence: for 
functions $f$ in some smoothness class, $f(x_n^*)$ should tend to $\min f$, 
preferably at some fast rate.  The simplest method would be to fix a sequence 
of $x_n$ in advance, and set $x^*_n = \arg \min \hat{f}_n$, for some 
approximation $\hat{f}_n$ to $f$. We will show that if $\hat{f}_n$ converges 
in supremum norm at the optimal rate, then $f(x_n^*)$ also converges at its 
optimal rate.  However, while this strategy gives a good worst-case bound, on 
average it is clearly a poor method of optimization: the design points $x_n$ 
are completely independent of the observations $z_n$.

We may therefore ask if there are more efficient methods, with better 
average-case performance, that nevertheless provide good guarantees of 
convergence.  The difficulty in designing such a method lies in the trade-off 
between {\em exploration} and {\em exploitation}.  If we exploit the data, 
observing in regions where $f$ is known to be low, we will be more likely to 
find the optimum quickly; however, unless we explore every region of $X$, we 
may not find it at all \citep{macready_bandit_1998}.

Initial attempts at this problem include work on Lipschitz optimization 
\citep[summarized in][]{hansen_global_1992} and the DIRECT algorithm  
\citep{jones_lipschitzian_1993}, but perhaps the best-known strategy is 
expected improvement.  It is sometimes called Bayesian optimization, and first 
appeared in \citet{mokus_bayesian_1974} as a Bayesian decision-theoretic 
solution to the problem.  Contemporary computers were not powerful enough to 
implement the technique in full, and it was later popularized by 
\citet{jones_efficient_1998}, who provided a computationally efficient 
implementation.  More recently, it has also been called a knowledge-gradient 
policy by \citet{frazier_knowledge-gradient_2009}.  Many extensions and 
alterations have been suggested by further authors; a good summary can be 
found in \citet{brochu_tutorial_2010}.

Expected improvement performs well in experiments 
\citep[\S9.5]{osborne_bayesian_2010}, but little is known about its 
theoretical properties.  The behaviour of the algorithm depends crucially on 
the Gaussian process prior $\pi$ chosen for $f$. Each prior has an associated 
space of functions $\mathcal{H}$, its reproducing-kernel Hilbert space. 
$\mathcal{H}$ contains all functions $X \to \R$ as smooth as a posterior mean 
of $f$, and is the natural space in which to study questions of convergence.

\citet{vazquez_convergence_2010} show that when $\pi$ is a fixed Gaussian 
process prior of finite smoothness, expected improvement converges on the 
minimum of any $f \in \mathcal{H}$, and almost surely for $f$ drawn from 
$\pi$.  \citet{grunewalder_regret_2010} bound the convergence rate of a 
computationally infeasible version of expected improvement: for priors $\pi$ 
of smoothness $\nu$, they show convergence at a rate $O^*(n^{-(\nu \wedge 
0.5)/d})$ on $f$ drawn from $\pi$.  We begin by bounding the convergence rate 
of the feasible algorithm, and show convergence at a rate $O^*(n^{-(\nu \wedge 
1)/d})$ on all $f \in \mathcal{H}$.  We go on to show that a modification of 
expected improvement converges at the near-optimal rate $O^*(n^{-\nu/d})$.

For practitioners, however, these results are somewhat misleading. In typical 
applications, the prior is not held fixed, but depends on parameters estimated 
sequentially from the data. This process ensures the choice of observations is 
invariant under translation and scaling of $f$, and is believed to be more 
efficient \citep[\S2]{jones_efficient_1998}. It has a profound effect on 
convergence, however: \citet[\S3.2]{locatelli_bayesian_1997} shows that, for a 
Brownian motion prior with estimated parameters, expected improvement may not 
converge at all.

We extend this result to more general settings, showing that for standard 
priors with estimated parameters, there exist smooth functions $f$ on which 
expected improvement does not converge.  We then propose alternative estimates 
of the prior parameters, chosen to minimize the constants in the convergence 
rate.  We show that these estimators give an automatic choice of parameters, 
while retaining the convergence rates of a fixed prior.

\autoref{tab:notation} summarizes the notation used in this paper.  We say $f: 
\R^d \to \R$ is a bump function if $f$ is infinitely differentiable and of 
compact support, and $f : \R^d \to \C$ is Hermitian if $\overline{f(x)} = 
f(-x)$.  We use the Landau notation $f = O(g)$ to denote $\lim \sup \abs{f/g} 
< \infty$, and $f = o(g)$ to denote $f/g \to 0$. If $g = O(f)$, we say $f = 
\Omega(g)$, and if both $f = O(g)$ and $f = \Omega(g)$, we say $f = 
\Theta(g)$. If further $f/g \to 1$, we say $f \sim g$. Finally, if $f$ and $g$ 
are random, and $\P(\sup \abs{f/g} \le M) \to 1$ as $M \to \infty$, we say $f 
= O_p(g)$.

In \autoref{sec:expected-improvement-algorithms}, we briefly describe the 
expected-improvement algorithm, and detail our assumptions on the priors used.  
We state our main results in \autoref{sec:convergence-rates}, and discuss 
implications for further work in \autoref{sec:discussion}. Finally, we give 
proofs in \autoref{sec:proofs}.

\begin{table}
  \begin{center}
    \begin{tabular}{cl}
      \toprule
      \multicolumn{2}{l}{\autoref{sec:introduction}} \\
      \midrule
      $f$ & unknown function $X \to \R$ to be minimized \\
      $X$ & compact subset of $\R^d$ to minimize over \\
      $d$ & number of dimensions to minimize over \\
      $x_n$ & points in $X$ at which $f$ is observed \\
      $z_n$ & observations $z_n = f(x_n)$ of $f$ \\
      $x_n^*$ & estimated minimum of $f$, given $z_1,\dots,z_n$ \\
      \midrule
      \multicolumn{2}{l}{\autoref{sec:bayesian-optimization}} \\
      \midrule
      $\pi$ & prior distribution for $f$ \\
      $u$ & strategy for choosing $x_n$, $x_n^*$\\
      $\mathcal F_n$ & filtration $\mathcal F_n = \sigma(x_i, z_i : i \le 
      n)$\\
      $z_n^*$ & best observation $z_n^* = \min_{i=1,\ldots,n}z_i$\\
      $EI_n$ & expected improvement given $\mathcal F_n$\\
      \midrule
      \multicolumn{2}{l}{\autoref{sec:gaussian-process-models}}\\
      \midrule
      $\mu$, $\sigma^2$ & global mean and variance of Gaussian-process prior 
      $\pi$ \\
      $K$ & underlying correlation kernel for $\pi$ \\
      $K_\theta$ & correlation kernel for $\pi$ with length-scales $\theta$ \\
      $\nu$, $\alpha$ & smoothness parameters of $K$ \\
      $\hat \mu_n$, $\hat f_n$, $s_n^2$, $\hat R_n^2$ & quantities describing 
      posterior distribution of $f$ given $\mathcal F_n$ \\
      \midrule
      \multicolumn{2}{l}{\autoref{sec:expected-improvement-strategies}}\\
      \midrule
      $EI(\pi)$ & expected improvement strategy with fixed prior \\
      $\hat \sigma^2_n$, $\hat \theta_n$ & estimates of prior parameters 
      $\sigma^2$, $\theta$ \\
      $c_n$ & rate of decay of $\hat \sigma_n^2$ \\
      $\theta^L$, $\theta^U$ & bounds on $\hat \theta_n$ \\
      $EI(\hat \pi)$ & expected improvement strategy with estimated prior \\
      \midrule
      \multicolumn{2}{l}{\autoref{sec:reproducing-kernel-hilbert-spaces}}\\
      \midrule
      $\mathcal H_\theta(S)$ & reproducing-kernel Hilbert space of $K_\theta$ 
      on $S$ \\
      $H^s(D)$ & Sobolev Hilbert space of order $s$ on $D$\\
      \midrule
      \multicolumn{2}{l}{\autoref{sec:fixed-parameters}}\\
      \midrule
      $L_n$ & loss suffered over an RKHS ball after $n$ steps\\
      \midrule
      \multicolumn{2}{l}{\autoref{sec:estimated-parameters}}\\
      \midrule
      $EI(\tilde \pi)$ & expected improvement strategy with robust estimated 
      prior\\
      \midrule
      \multicolumn{2}{l}{\autoref{sec:optimal-rates}}\\
      \midrule
      $EI(\,\cdot\,, \varepsilon)$ & $\varepsilon$-greedy expected improvement 
      strategies\\
      \bottomrule
    \end{tabular}
  \end{center}
  \caption{Notation}
  \label{tab:notation}
\end{table}

\section{Expected Improvement}
\label{sec:expected-improvement-algorithms}

Suppose we wish to minimize an unknown function $f$, choosing design points 
$x_n$ and estimated minima $x_n^*$ as in the introduction. If we pick a prior 
distribution $\pi$ for $f$, representing our beliefs about the unknown 
function, we can describe this problem in terms of decision theory.  Let 
$(\Omega, \mathcal{F}, \P)$ be a probability space, equipped with a random 
process $f$ having law $\pi$. A strategy $u$ is a collection of random 
variables $(x_n)$, $(x_n^*)$ taking values in $X$.  Set $z_n \coloneqq f(x_n)$, and 
define the filtration $\mathcal{F}_n \coloneqq \sigma(x_i, z_i : i \le n)$. The 
strategy $u$ is valid if $x_n$ is conditionally independent of $f$ given 
$\mathcal{F}_{n-1}$, and likewise $x_n^*$ given $\mathcal{F}_n$.  (Note that 
we allow random strategies, provided they do not depend on unknown information 
about $f$.)

When taking probabilities and expectations we will write $\P^u_\pi$ and 
$\E^u_\pi$, denoting the dependence on both the prior $\pi$ and strategy $u$.  
The average-case performance at some future time $N$ is then given by the 
expected loss,
\[\E^u_\pi[f(x_N^*) - \min f],\]
and our goal, given $\pi$, is to choose the strategy $u$ to minimize this 
quantity.

\subsection{Bayesian Optimization}
\label{sec:bayesian-optimization}

For $N > 1$ this problem is very computationally intensive 
\citep[\S6.3]{osborne_bayesian_2010}, but we can solve a simplified version of 
it.  First, we restrict the choice of $x_n^*$ to the previous design points 
$x_1, \dots, x_n$.  (In practice this is reasonable, as choosing an $x_n^*$ we 
have not observed can be unreliable.)  Secondly, rather than finding an 
optimal strategy for the problem, we derive the myopic strategy: the strategy 
which is optimal if we always assume we will stop after the next observation.  
This strategy is suboptimal \citep[\S3.1]{ginsbourger_multi-points_2008}, but 
performs well, and greatly simplifies the calculations involved.

In this setting, given $\mathcal{F}_n$, if we are to stop at time $n$ we 
should choose $x^*_n \coloneqq x_{i_n^*}$, where $i_n^* \coloneqq \arg 
\min_{1, \dots, n} z_i$. (In the case of ties, we may pick any minimizing 
$i_n^*$.) We then suffer a loss $z_n^* - \min f$, where $z_n^* \coloneqq 
z_{i_n^*}$.  Were we to observe at $x_{n+1}$ before stopping, the expected 
loss would be
\[\E_\pi^u[z_{n+1}^* - \min f\mid \mathcal{F}_n],\]
so the myopic strategy should choose $x_{n+1}$ to minimize this quantity.  
Equivalently, it should maximize the expected improvement over the current 
loss,
\begin{equation}
  \label{eq:expected-improvement}
  EI_n(x_{n+1}; \pi) \coloneqq \E_\pi^u[z^*_n - z_{n+1}^* \mid \mathcal{F}_n] 
  = \E_\pi^u[(z^*_n - z_{n+1})^+\mid \mathcal{F}_n],
\end{equation}
where $x^+ = \max(x, 0)$.

So far, we have merely replaced one optimization problem with another.  
However, for suitable priors, $EI_n$ can be evaluated cheaply, and thus
maximized by standard techniques. The expected-improvement algorithm is then 
given by choosing $x_{n+1}$ to maximize \eqref{eq:expected-improvement}.

\subsection{Gaussian Process Models}
\label{sec:gaussian-process-models}

We still need to choose a prior $\pi$ for $f$. Typically, we model $f$ as a 
stationary Gaussian process: we consider the values $f(x)$ to be jointly 
Gaussian, with mean and covariance
\begin{equation}
  \label{eq:gaussian-process-defn}
  \E_\pi[f(x)] = \mu, \quad \Cov_\pi[f(x), f(y)] = \sigma^2 K_\theta(x-y).
\end{equation}
$\mu \in \R$ is the global mean of $f$; we place a flat prior on $\mu$, 
reflecting our uncertainty over the location of $f$.

$\sigma > 0$ is the global scale of variation of $f$, and $K_\theta : \R^d \to 
\R$ its correlation kernel, governing the local properties of $f$.  In the 
following, we will consider kernels
\begin{equation}
  \label{eq:kernel-defn}
  K_\theta(t_1, \dots, t_d) \coloneqq K(t_1/\theta_1,\dots,t_d/\theta_d),
\end{equation}
for an underlying kernel $K$ with $K(0) = 1$. (Note that we can always satisfy 
this condition by suitably scaling $K$ and $\sigma$.) The $\theta_i > 0$ are 
the length-scales of the process: two values $f(x)$ and $f(y)$ will be highly 
correlated if each $x_i - y_i$ is small compared with $\theta_i$.  For now, we 
will assume the parameters $\sigma$ and $\theta$ are fixed in advance.

For \eqref{eq:gaussian-process-defn} and \eqref{eq:kernel-defn} to define a 
consistent Gaussian process, $K$ must be a symmetric positive-definite 
function.  We will also make the following assumptions.

\begin{assumption}
  \label{ass:1}
$K$ is continuous and integrable.
\end{assumption}
\noindent
$K$ thus has Fourier transform
\[\widehat{K}(\xi) \coloneqq \int_{\R^d} e^{-2\pi i\langle x, \xi\rangle} K(x) 
\, dx,\]
and by Bochner's theorem, $\widehat{K}$ is non-negative and integrable.

\begin{assumption}
$\widehat{K}$ is isotropic and radially non-increasing.
\end{assumption}
\noindent
In other words, $\widehat{K}(x) = \widehat{k}(\norm{x})$ for a non-increasing 
function $\widehat{k} : [0, \infty) \to [0, \infty)$; as a consequence, $K$ is 
isotropic.

\begin{assumption}
  As $x \to \infty$, either:
\begin{enumerate}
  \item $\widehat{K}(x) = \Theta(\norm{x}^{-2\nu - d})$ for some $\nu > 0$; or
  \item $\widehat{K}(x) = O(\norm{x}^{-2\nu - d})$ for all $\nu > 0$ (we will 
    then say that $\nu = \infty$).
\end{enumerate}
\end{assumption}
\noindent
Note the condition $\nu > 0$ is required for $\widehat{K}$ to be integrable.

\begin{assumption}
  \label{ass:4}
  $K$ is $C^k$, for $k$ the largest integer less than $2\nu$, and at the 
  origin, $K$ has $k$-th order Taylor approximation $P_k$ satisfying
  \[\abs{K(x) - P_k(x)} = O\left(\norm{x}^{2\nu}(-\log 
  \norm{x})^{2\alpha}\right)\]
  as $x \to 0$, for some $\alpha \ge 0$.
\end{assumption}
\noindent
When $\alpha = 0$, this is just the condition that $K$ be $2\nu$-H\"{o}lder at 
the origin; when $\alpha > 0$, we instead require this condition up to a log 
factor.

The rate $\nu$ controls the smoothness of functions from the prior: almost 
surely, $f$ has continuous derivatives of any order $k < \nu$ 
\citep[\S1.4.2]{adler_random_2007}.  Popular kernels include the Mat\'{e}rn 
class,
\[K^\nu(x) \coloneqq 
\frac{2^{1-\nu}}{\Gamma(\nu)}\left(\sqrt2\nu\norm{x}\right)^\nu 
k_\nu\left(\sqrt2\nu\norm{x}\right), \quad \nu \in (0, \infty),\]
where $k_\nu$ is a modified Bessel function of the second kind, and the 
Gaussian kernel,
\[K^\infty(x) \coloneqq e^{-\frac12 \norm{x}^2},\]
obtained in the limit $\nu \to \infty$ \citep[\S4.2]{rasmussen_gaussian_2006}.  
Between them, these kernels cover the full range of smoothness $0 < \nu \le 
\infty$.  Both kernels satisfy 
Assumptions~\ref{ass:1}--\ref{ass:4} for the $\nu$ given; 
$\alpha = 0$ except for the Mat\'{e}rn kernel with $\nu \in \N$, where $\alpha 
= \frac12$ \citep[\S9.6]{abramowitz_handbook_1965}.

Having chosen our prior distribution, we may now derive its posterior.  We 
find
\[f(x) \mid z_1, \dots, z_{n} \sim N\left(\hat{f}_n(x; \theta), \sigma^2 
s_n^2(x; \theta)\right),\]
where
\begin{align}
  \label{eq:pred-mu}
  \hat{\mu}_n(\theta) &\coloneqq \frac{1^TV^{-1}z}{1^TV^{-1}1},\\
  \label{eq:pred-mean}
  \hat{f}_n(x; \theta) &\coloneqq \hat{\mu}_n + v^TV^{-1}(z - \hat{\mu}_n1),\\  
  \shortintertext{and}
  \label{eq:pred-var}
  s^2_n(x; \theta) &\coloneqq 1 - v^TV^{-1}v + \frac{(1 - 
  1^TV^{-1}v)^2}{1^TV^{-1}1},
\end{align}
for $z = (z_i)_{i=1}^{n}$, $V = (K_\theta(x_i-x_j))_{i,j=1}^{n}$, and $v = 
(K_\theta(x - x_i))_{i=1}^{n}$ \citep[\S4.1.3]{santner_design_2003}.  
Equivalently, these expressions are the best linear unbiased predictor of 
$f(x)$ and its variance, as given in \citet[\S2]{jones_efficient_1998}. We 
will also need the reduced sum of squares,
\begin{equation}
  \label{eq:rss}
  \hat{R}_n^2(\theta) \coloneqq (z - \hat{\mu}_n1)^T V^{-1} (z-\hat{\mu}_n1).
\end{equation}

\subsection{Expected Improvement Strategies}
\label{sec:expected-improvement-strategies}

Under our assumptions on $\pi$, we may now derive an analytic form for 
\eqref{eq:expected-improvement},
as in \citet[\S4.1]{jones_efficient_1998}. We obtain
\begin{equation}
  \label{eq:ei-formula}
  EI_n(x_{n+1}; \pi) = \rho\left(z^*_n - \hat{f}_n(x_{n+1}; \theta), \sigma 
  s_n(x_{n+1}; \theta)\right),
\end{equation}
where
\begin{equation}
  \label{eq:ei-rho}
\rho(y, s) \coloneqq \begin{cases} y\Phi(y/s) + s\varphi(y/s), & s > 0, \\ 
  \max(y, 0), & s = 0, \end{cases}
\end{equation}
and $\Phi$ and $\varphi$ are the standard normal distribution and density 
functions respectively.

For a prior $\pi$ as above, expected improvement chooses $x_{n+1}$ to maximize 
\eqref{eq:ei-formula}, but this does not fully define the strategy.  Firstly, 
we must describe how the strategy breaks ties, when more than one $x \in X$ 
maximizes $EI_n$. In general, this will not affect the behaviour of the 
algorithm, so we allow any choice of $x_{n+1}$ maximizing 
\eqref{eq:ei-formula}.

Secondly, we must say how to choose $x_1$, as the above expressions are 
undefined when $n=0$.  In fact, \citet[\S4.2]{jones_efficient_1998} find that 
expected improvement can be unreliable given few data points, and recommend 
that several initial design points be chosen in a random quasi-uniform 
arrangement.  We will therefore assume that until some fixed time $k$, points 
$x_1, \dots, x_k$ are instead chosen by some (potentially random) method 
independent of $f$. We thus obtain the following strategy.

\begin{definition}
  \label{def:ei}
  An $EI(\pi)$ strategy chooses:
  \begin{enumerate}
    \item initial design points $x_1, \dots, x_k$ independently of $f$; and
    \item further design points $x_{n+1}\ (n \ge k)$ from the maximizers of
      \eqref{eq:ei-formula}.
  \end{enumerate}
\end{definition}

So far, we have not considered the choice of parameters $\sigma$ and $\theta$.  
While these can be fixed in advance, doing so requires us to specify 
characteristic scales of the unknown function $f$, and causes expected 
improvement to behave differently on a rescaling of the same function. We 
would prefer an algorithm which could adapt automatically to the scale of $f$.

A natural approach is to take maximum likelihood estimates of the parameters, 
as recommended by \citet[\S2]{jones_efficient_1998}. Given $\theta$, the MLE  
$\hat{\sigma}^2_n = \hat{R}_n^2(\theta)/n$; for full generality, we will allow 
any choice $\hat \sigma^2_n = c_n \hat R_n^2(\theta)$, where $c_n = o(1/\log 
n)$.  Estimates of $\theta$, however, must be obtained by numerical 
optimization. As $\theta$ can vary widely in scale, this optimization is best 
performed over $\log \theta$; as the likelihood surface is typically 
multimodal, this requires the use of a global optimizer.  We must therefore 
place (implicit or explicit) bounds on the allowed values of $\log \theta$. We 
have thus described the following strategy.

\begin{definition}
  \label{def:ei-hat}
  Let $\hat \pi_n$ be a sequence of priors, with parameters $\hat \sigma_n$, 
  $\hat \theta_n$ satisfying:
  \begin{enumerate}
    \item $\hat \sigma^2_n = c_n \hat R_n^2(\hat \theta_n)$ for constants $c_n 
      > 0$, $c_n = o(1/\log n)$; and
    \item $\theta^L \le \hat \theta_n \le \theta^U$ for constants $\theta^L$,  
      $\theta^U \in \R_+^d$.
  \end{enumerate}
  An $EI(\hat \pi)$ strategy satisfies \autoref{def:ei}, replacing $\pi$ with 
  $\hat \pi_n$ in \eqref{eq:ei-formula}.
\end{definition}

\section{Convergence Rates}
\label{sec:convergence-rates}

To discuss convergence, we must first choose a smoothness class for the 
unknown function $f$. Each kernel $K_\theta$ is associated with a space of 
functions $\mathcal{H}_\theta(X)$, its reproducing-kernel Hilbert space (RKHS) 
or native space. $\mathcal H_\theta(X)$ contains all functions $X \to \R$ as 
smooth as a posterior mean of $f$, and is the natural space to study 
convergence of expected-improvement algorithms, allowing a tractable analysis 
of their asymptotic behaviour.

\subsection{Reproducing-Kernel Hilbert Spaces}
\label{sec:reproducing-kernel-hilbert-spaces}

Given a symmetric positive-definite kernel $K$ on $\R^d$, set $k_x(t) = K(t - 
x)$. For $S \subseteq \R^d$, let $\mathcal{E}(S)$ be the space of functions $S 
\to \R$ spanned by the $k_x$, for $x \in S$.  Furnish $\mathcal{E}(S)$ with 
the inner product defined by
\[\langle k_x, k_y\rangle \coloneqq K(x - y).\]
The completion of $\mathcal{E}(S)$ under this inner product is the 
reproducing-kernel Hilbert space $\mathcal{H}(S)$ of $K$ on $S$. The members 
$f \in \mathcal{H}(S)$ are abstract objects, but we can identify them with 
functions $f : S \to \R$ through the reproducing property,
\[f(x) = \langle f,k_x\rangle,\]
which holds for all $f \in \mathcal{E}(S)$. See \citet{aronszajn_theory_1950}, 
\citet{berlinet_reproducing_2004}, \citet{wendland_scattered_2005} and 
\citet{van_der_vaart_reproducing_2008}.

We will find it convenient also to use an alternative characterization of 
$\mathcal H(S)$.  We begin by describing $\mathcal{H}(\R^d)$ in terms of 
Fourier transforms.  Let $\widehat f$ denote the Fourier transform of a 
function $f \in L^2$.  The following result is stated in 
\citet[\S2]{parzen_probability_1963}, and proved in 
\citet[\S10.2]{wendland_scattered_2005}; we give a short proof in 
\autoref{sec:proofs}.

\begin{lemma}
  \label{lem:rkhs-fourier}
  $\mathcal{H}(\R^d)$ is the space of real continuous $f \in L^2(\R^d)$ whose 
  norm
  \[\norm{f}^2_{\mathcal{H}(\R^d)} \coloneqq \int 
  \frac{\abs{\widehat{f}(\xi)}^2}{\widehat{K}(\xi)} \, d\xi\]
  is finite, taking $0/0 = 0$.
\end{lemma}

We may now describe $\mathcal{H}(S)$ in terms of $\mathcal{H}(\R^d)$.

\begin{lemma}[\citealp{aronszajn_theory_1950}, \S1.5]
  $\mathcal{H}(S)$ is the space of functions $f = g|_S$ for some
  $g \in \mathcal{H}(\R^d)$, with norm \[\norm{f}_{\mathcal{H}(S)} \coloneqq 
  \inf_{g|_S = f} \norm{g}_{\mathcal{H}(\R^d)},\] and there is a unique $g$ 
  minimizing this expression.
  \label{lem:rkhs-restriction}
\end{lemma}

These spaces are in fact closely related to the Sobolev Hilbert spaces of 
functional analysis. Say a domain $D \subseteq \R^d$ is Lipschitz if its 
boundary is locally the graph of a Lipschitz function (see 
\citealp[\S12]{tartar_introduction_2007}, for a precise definition). For such 
a domain $D$, the Sobolev Hilbert space $H^s(D)$ is the space of functions $f 
: D \to \R$, given by the restriction of some $g : \R^d \to \R$, whose norm
\[\norm{f}^2_{H^s(D)} \coloneqq \inf_{g|_D=f} \int \frac{\abs{\widehat 
g(\xi)}^2}{(1+\norm{\xi}^2)^{s/2}}\, d\xi\]
is finite.  Thus, for the kernel $K$ with Fourier transform $\widehat K(\xi) = 
(1 + \norm{\xi}^2)^{s/2}$, this is just the RKHS $\mathcal H(D)$. More generally, 
if $K$ satisfies our assumptions with $\nu < \infty$, these spaces are 
equivalent in the sense of normed spaces: they contain the same functions, and 
have norms $\norm{\,\cdot\,}_1, \norm{\,\cdot\,}_2$ satisfying
\[C\norm{f}_1 \le \norm{f}_2 \le C'\norm{f}_1,\]
for constants $0 < C \le C'$.

\begin{lemma}
  \label{lem:rkhs-embedding}
  Let $\mathcal H_\theta(S)$ denote the RKHS of $K_\theta$ on $S$, and $D 
  \subseteq \R^d$ be a Lipschitz domain.
  \begin{enumerate}
    \item If $\nu < \infty$, $\mathcal{H}_\theta(\bar{D})$ is equivalent to 
      the Sobolev Hilbert space $H^{\nu+d/2}(D)$.
    \item If $\nu = \infty$, $\mathcal{H}_\theta(\bar{D})$ is continuously 
      embedded in $H^s(D)$ for all $s$.
  \end{enumerate}
\end{lemma}

Thus if $\nu < \infty$, and $X$ is, say, a product of intervals $\prod_{i=1}^d 
[a_i, b_i]$, the RKHS $\mathcal{H}_\theta(X)$ is equivalent to the Sobolev 
Hilbert space $H^{\nu+d/2}(\prod_{i=1}^d(a_i, b_i))$, identifying each 
function in that space with its unique continuous extension to $X$.

\subsection{Fixed Parameters}
\label{sec:fixed-parameters}

We are now ready to state our main results. Let $X \subset \R^d$ be compact 
with non-empty interior. For a function $f :X \to \R$, let $\P_f^u$ and 
$\E_f^u$ denote probability and expectation when minimizing the fixed function 
$f$ with strategy $u$.  (Note that while $f$ is fixed, $u$ may be random, so 
its performance is still probabilistic in nature.) We define the loss suffered 
over the ball $B_R$ in $\mathcal{H}_\theta(X)$ after $n$ steps by a strategy 
$u$,
\[L_n(u, \mathcal{H}_\theta(X), R) \coloneqq 
\sup_{\norm{f}_{\mathcal{H}_\theta(X)} \le R} \E_f^u[f(x_n^*)-\min f].\]
We will say that $u$ converges on the optimum at rate $r_n$, if
\[L_n(u, \mathcal{H}_\theta(X), R) = O(r_n)\]
for all $R > 0$. Note that we do not allow $u$ to vary with $R$; the strategy 
must achieve this rate without prior knowledge of 
$\norm{f}_{\mathcal{H}_\theta(X)}$.

We begin by showing that the minimax rate of convergence is $n^{-\nu/d}$.  

\begin{theorem}
  \label{thm:minimax-rates}
  If $\nu < \infty$, then for any $\theta \in \R^d_+$, $R > 0$,
  \[\inf_u L_n(u, \mathcal{H}_\theta(X), R) = \Theta(n^{-\nu/d}),\]
  and this rate can be achieved by a strategy $u$ not depending on $R$.
\end{theorem}

The upper bound is provided by a naive strategy as in the introduction: we fix 
a quasi-uniform sequence $x_n$ in advance, and take $x_n^*$ to minimize a 
radial basis function interpolant of the data.  As remarked previously, 
however, this naive strategy is not very satisfying; in practice it will be 
outperformed by any good strategy varying with the data.  We may thus ask 
whether more sophisticated strategies, with better practical performance, can 
still provide good worst-case bounds.

One such strategy is the $EI(\pi)$ strategy of \autoref{def:ei}.  We can show 
this strategy converges at least at rate $n^{-(\nu \wedge 1)/d}$, up to log 
factors.

\begin{theorem}
  \label{thm:ei-rates}
  Let $\pi$ be a prior with length-scales $\theta \in \R^d_+$. For any $R > 
  0$,
  \[L_n(EI(\pi), \mathcal{H}_\theta(X), R) = \begin{cases}
    O(n^{-\nu/d}(\log n)^{\alpha}), & \nu \le 1,\\
    O(n^{-1/d}), & \nu > 1.\\
  \end{cases} \]
\end{theorem}

For $\nu \le 1$, these rates are near-optimal. For $\nu > 1$, we are faced 
with a more difficult problem; we discuss this in more detail in 
\autoref{sec:optimal-rates}.

\subsection{Estimated Parameters}
\label{sec:estimated-parameters}

First, we consider the effect of the prior parameters on $EI(\pi)$.  While the 
previous result gives a convergence rate for any fixed choice of parameters, 
the constant in that rate will depend on the parameters chosen; to choose 
well, we must somehow estimate these parameters from the data.  The 
$EI(\hat{\pi})$ strategy, given by \autoref{def:ei-hat}, uses maximum 
likelihood estimates for this purpose. We can show, however, that this may 
cause the strategy to never converge.

\begin{theorem}
  \label{thm:ei-hat-diverges}
  Suppose $\nu < \infty$. Given $\theta \in \R^d_+$, $R > 0$, $\varepsilon > 
  0$, there exists $f \in \mathcal{H}_\theta(X)$ satisfying 
  $\norm{f}_{\mathcal{H}_\theta(X)} \le R$, and for some fixed $\delta > 0$,
  \[\P_f^{EI(\hat{\pi})}\left(\inf_n f(x_n^*) - \min f \ge \delta\right) > 1 - 
  \varepsilon.\]
\end{theorem}

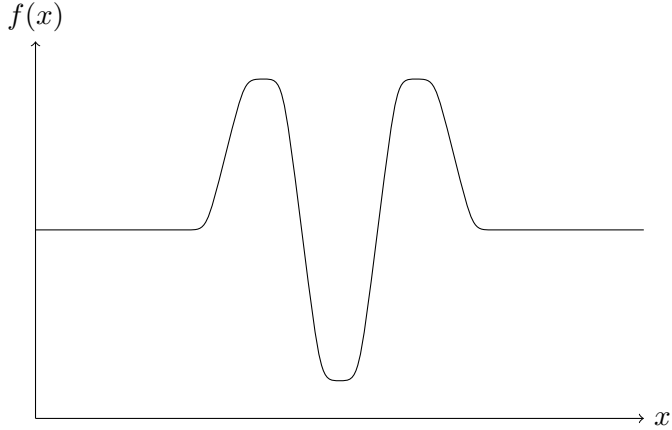
\begin{figure}
  \begin{center}
    \begin{tikzpicture}
      \draw[->] (1,-2.5) -- (9,-2.5) node[right] {$x$};
      \draw[->] (1,-2.5) -- (1,2.5) node[above] {$f(x)$};
      \draw (1,0) -- (3,0) .. controls (3.25,0) .. (3.5,1) ..  controls 
      (3.75,2) .. (4,2) .. controls (4.25,2) .. (4.5,0) ..  controls (4.75,-2) 
      ..  (5,-2) .. controls (5.25,-2) .. (5.5,0) .. controls (5.75,2) .. 
      (6,2) ..  controls (6.25, 2) .. (6.5, 1) .. controls (6.75, 0) .. (7, 0) 
      -- (9, 0);
    \end{tikzpicture}
  \end{center}
  \caption{A counterexample from \autoref{thm:ei-hat-diverges}}
  \label{fig:counterexample-fn}
\end{figure}

The counterexamples constructed in the proof of the theorem may be 
difficult to minimize, but they are not badly-behaved 
(\autoref{fig:counterexample-fn}). A good optimization strategy should be able 
to minimize such functions, and we must ask why expected improvement fails.

We can understand the issue by considering the constant in  
\autoref{thm:ei-rates}.  Define
\begin{equation*}
  \tau(x) \coloneqq x \Phi(x) + \varphi(x).
\end{equation*}
From the proof of \autoref{thm:ei-rates}, the dominant term in the convergence 
rate has constant
\begin{equation}
  C(R + \sigma)\frac{\tau(R/\sigma)}{\tau(-R/\sigma)}  
  \label{eq:rate-constant},
\end{equation}
for $C > 0$ not depending on $R$ or $\sigma$. In \autoref{sec:proofs}, we will 
prove the following result.
\begin{corollary}
  \label{cor:r-hat}
  $\hat{R}_n(\theta)$ is non-decreasing in $n$, and bounded above by 
  $\norm{f}_{\mathcal{H}_\theta(X)}$.
\end{corollary}
\noindent
Hence for fixed $\theta$, the estimate $\hat{\sigma}_n^2 = 
\hat{R}_n^2(\theta)/n \le R^2/n$, and thus $R/\hat{\sigma}_n \ge n^{1/2}$.  
Inserting this choice into \eqref{eq:rate-constant} gives a constant growing 
exponentially in $n$, destroying our convergence rate.

To resolve the issue, we will instead try to pick $\sigma$ to minimize 
\eqref{eq:rate-constant}.  The term $R + \sigma$ is increasing in $\sigma$, 
and the term $\tau(R/\sigma) / \tau(-R/\sigma)$ is decreasing in $\sigma$; we 
may balance the terms by taking $\sigma = R$. The constant is then 
proportional to $R$, which we may minimize by taking $R = 
\norm{f}_{\mathcal{H}_\theta(X)}$.  In practice, we will not know 
$\norm{f}_{\mathcal{H}_\theta(X)}$ in advance, so we must estimate it from the 
data; from \autoref{cor:r-hat}, a convenient estimate is $\hat{R}_n(\theta)$.

Suppose, then, that we make some bounded estimate $\hat \theta_n$ of $\theta$, 
and set $\hat{\sigma}_n^2 = \hat{R}_n^2(\hat{\theta}_n)$. As 
\autoref{thm:ei-hat-diverges} holds for any $\hat \sigma^2_n$ of faster than 
logarithmic decay, such a choice is necessary to ensure convergence.
(We may also choose $\theta$ to minimize \eqref{eq:rate-constant}; we might 
then pick $\hat{\theta}_n$ minimizing $\hat{R}_n(\theta)\prod_{i=1}^d 
\theta_i^{-\nu/d},$
but our assumptions on $\hat{\theta}_n$ are weak enough that we need not 
consider this further.)

If we believe our Gaussian-process model, this estimate $\hat \sigma_n$ is 
certainly unusual. We should, however, take care before placing too much faith 
in the model. The function in \autoref{fig:counterexample-fn} is a reasonable 
function to optimize, but as a Gaussian process it is highly atypical: there 
are intervals on which the function is constant, an event which in our model 
occurs with probability zero. If we want our algorithm to succeed on more 
general classes of functions, we will need to choose our parameter estimates 
appropriately.

To obtain good rates, we must add a further condition to our strategy.  If 
$z_1 = \dots = z_n$, $EI_n(\,\cdot\,; \hat{\pi}_n)$ is identically zero, and 
all choices of $x_{n+1}$ are equally valid.  To ensure we fully explore $f$, 
we will therefore require that when our strategy is applied to a constant 
function $f(x) = c$, it produces a sequence $x_n$ dense in $X$. (This can be 
achieved, for example, by choosing $x_{n+1}$ uniformly at random from $X$ when 
$z_1 = \dots = z_n$.) We have thus described the following strategy.

\begin{definition}
  \label{def:ei-tilde}
  An $EI(\tilde \pi)$ strategy satisfies \autoref{def:ei-hat}, except:
  \begin{enumerate}
    \item we instead set $\hat \sigma_n^2 = \hat R_n^2(\hat \theta_n)$; and
    \item we require the choice of $x_{n+1}$ maximizing \eqref{eq:ei-formula} 
      to be such that, if $f$ is constant, the design points are almost surely 
      dense in $X$.
  \end{enumerate}
\end{definition}

We cannot now prove a convergence result uniform over balls in 
$\mathcal{H}_\theta(X)$, as the rate of convergence depends on the ratio 
$R/\hat{R}_n$, which is unbounded. (Indeed, any estimator of 
$\norm{f}_{\mathcal{H}_\theta(X)}$ must sometimes perform poorly: $f$ can appear 
from the data to have arbitrarily small norm, while in fact having a spike 
somewhere we have not yet observed.) We can, however, provide the same 
convergence rates as in \autoref{thm:ei-rates}, in a slightly weaker sense.

\begin{theorem}
  \label{thm:ei-tilde-rates}
  For any $f \in \mathcal{H}_{\theta^U}(X)$, under $\P_f^{EI(\tilde\pi)}$,
  \[f(x_n^*) - \min f = \begin{cases}
    O_p(n^{-\nu/d}(\log n)^{\alpha}), & \nu \le 1,\\
    O_p(n^{-1/d}), & \nu > 1.
  \end{cases} \]
\end{theorem}

\subsection{Near-Optimal Rates}
\label{sec:optimal-rates}

So far, our rates have been near-optimal only for $\nu \le 1$. To obtain good 
rates for $\nu > 1$, standard results on the performance of Gaussian-process 
interpolation \citep[\S6]{narcowich_refined_2003} then require the design 
points $x_i$ to be quasi-uniform in a region of interest. It is unclear 
whether this occurs naturally under expected improvement, but there are many 
ways we can modify the algorithm to ensure it.

Perhaps the simplest, and most well-known, is an $\varepsilon$-greedy strategy 
\citep[\S2.2]{sutton_reinforcement_1998}. In such a strategy, at each step 
with probability $1 - \varepsilon$ we make a decision to maximize some greedy 
criterion; with probability $\varepsilon$ we make a decision completely at 
random. This random choice ensures that the short-term nature of the greedy 
criterion does not overshadow our long-term goal.

The parameter $\varepsilon$ controls the trade-off between global and local 
search: a good choice of $\varepsilon$ will be small enough to not interfere 
with the expected-improvement algorithm, but large enough to prevent it from 
getting stuck in a local minimum.  \citet[\S2.2]{sutton_reinforcement_1998} 
consider the values $\varepsilon = 0.1$ and $\varepsilon = 0.01$, but in 
practical work $\varepsilon$ should of course be calibrated to a typical 
problem set.

We therefore define the following strategies.
\begin{definition}
  \label{def:ei-eps}
  Let $\cdot$ denote $\pi$, $\hat \pi$ or $\tilde \pi$. For $0 < \varepsilon < 
  1$, an $EI(\,\cdot\,, \varepsilon)$ strategy:
  \begin{enumerate}
    \item chooses initial design points $x_1, \cdots, x_k$ independently of 
      $f$;
    \item with probability $1-\varepsilon$, chooses design point $x_{n+1}\ (n 
      \ge k)$ as in $EI(\,\cdot\,)$; or
    \item with probability $\varepsilon$, chooses $x_{n+1}\ (n \ge k)$ 
      uniformly at random from $X$.
  \end{enumerate}
\end{definition}

We can show that these strategies achieve near-optimal rates of convergence 
for all $\nu < \infty$.

\begin{theorem}
  \label{thm:ei-epsilon-optimal}

  Let $EI(\,\cdot\,, \varepsilon)$ be one of the strategies in 
  \autoref{def:ei-eps}.  If $\nu < \infty$, then for any $R > 0$,
  \[L_n(EI(\,\cdot\,, \varepsilon), \mathcal{H}_{\theta^U}(X), R) = O( (n/\log 
  n)^{-\nu/d}(\log n)^{\alpha}),\]
  while if $\nu = \infty$, the statement holds for all $\nu < \infty$.
\end{theorem}

Note that unlike a typical $\varepsilon$-greedy algorithm, we do not rely on 
random choice to obtain global convergence: as above, the $EI(\pi)$ and 
$EI(\tilde{\pi})$ strategies are already globally convergent.  Instead, we use 
random choice simply to improve upon the worst-case rate.  Note also that the 
result does not in general hold when $\varepsilon = 1$; to obtain good rates, 
we must combine global search with inference about $f$.

\section{Conclusions}
\label{sec:discussion}

We have shown that expected improvement can converge near-optimally, but a 
naive implementation may not converge at all. We thus echo 
\citet{diaconis_consistency_1986} in stating that, for infinite-dimensional 
problems, Bayesian methods are not always guaranteed to find the right answer; 
such guarantees can only be provided by considering the problem at hand.

We might ask, however, if our framework can also be improved. Our upper bounds 
on convergence were established using naive algorithms, which in practice
would prove inefficient. If a sophisticated algorithm fails where a naive one 
succeeds, then the sophisticated algorithm is certainly at fault; we might, 
however, prefer methods of evaluation which do not consider naive algorithms 
so successful.

\citet{vazquez_convergence_2010} and \citet{grunewalder_regret_2010} consider a 
more Bayesian formulation of the problem, where the unknown function $f$ is 
distributed according to the prior $\pi$, but this approach can prove 
restrictive: as we saw in \autoref{sec:estimated-parameters}, placing too much 
faith in the prior may exclude functions of interest.  Further, 
\citeauthor{grunewalder_regret_2010}\ find the same issues are present also 
within the Bayesian framework.

A more interesting approach is given by the continuum-armed-bandit problem 
\citep[and references therein]{srinivas_gaussian_2010}. Here the goal is to 
minimize the cumulative regret,
\[R_n \coloneqq \sum_{i=1}^n ( f(x_i) - \min f ),\]
in general observing the function $f$ under noise. Algorithms controlling the 
cumulative regret at rate $r_n$ also solve the optimization problem, at rate 
$r_n/n$ \citep[\S3]{bubeck_pure_2009}. The naive algorithms above, however, 
have poor cumulative regret.  We might, then, consider the cumulative regret 
to be a better measure of performance, but this approach too has limitations.  
Firstly, the cumulative regret is necessarily increasing, so cannot establish 
rates of optimization faster than $n^{-1}$. (This is not an issue under noise, 
where typically $r_n = \Omega(n^{1/2})$, see \citealp{kleinberg_sharp_2010}.) 
Secondly, if our goal is optimization, then minimizing the regret, a cost we 
do not incur, may obscure the problem at hand.

\citet{bubeck_x-armed_2010} study this problem with the additional assumption 
that $f$ has finitely many minima, and is, say, quadratic in a neighbourhood 
of each.  This assumption may suffice in practice, and allows the authors to 
obtain impressive rates of convergence. For optimization, however, a further 
weakness is that these rates hold only once the algorithm has found a basin of 
attraction; they thus measure local, rather than global, performance.  It may 
be that convergence rates alone are not sufficient to capture the performance 
of a global optimization algorithm, and the time taken to find a basin of 
attraction is more relevant. In any case, the choice of an appropriate 
framework to measure performance in global optimization merits further study.

Finally, we should also ask how to choose the smoothness parameter $\nu$ (or 
the equivalent parameter in similar algorithms).  
\Citet{van_der_vaart_adaptive_2009} show that Bayesian Gaussian-process models 
can, in some contexts, automatically adapt to the smoothness of an unknown 
function $f$. Their technique requires, however, that the estimated 
length-scales $\hat \theta_n$ to tend to 0, posing both practical and 
theoretical challenges.  The question of how best to optimize functions of 
unknown smoothness remains open.

\acks{We would like to thank the referees, as well as Richard Nickl and 
Steffen Grunewalder, for their valuable comments and suggestions.}

\appendix

\section{Proofs}
\label{sec:proofs}

We now prove the results in \autoref{sec:convergence-rates}.

\subsection{Reproducing-Kernel Hilbert Spaces}

\begin{proof}[Proof of \autoref{lem:rkhs-fourier}]
  Let $V$ be the space of functions described, and $W$ be the closed real 
  subspace of Hermitian functions in $L^2(\R^d, \widehat{K}^{-1})$. We will 
  show $f \mapsto \widehat f$ is an isomorphism $V \to W$, so we may 
  equivalently work with $W$. Given $\widehat{f} \in W$, by Cauchy-Schwarz and 
  Bochner's theorem,
  \[\int \abs{\widehat{f}} \le \left(\int \widehat{K}\right)^{1/2}
  \left(\int \abs{\widehat{f}}^2/\widehat{K}\right)^{1/2} < \infty,\]
  and as $\norm{\widehat{K}}_\infty \le \norm{K}_1$,
  \[\int \abs{\widehat{f}}^2 \le \norm{\widehat{K}}_{\infty} \int 
  \abs{\widehat{f}}^2/\widehat{K} < \infty,\]
  so $\widehat{f} \in L^1 \cap L^2$. $\widehat{f}$ is thus the Fourier 
  transform of a real continuous $f \in L^2$, satisfying the Fourier inversion 
  formula everywhere.

  $f \mapsto \widehat{f}$ is hence an isomorphism $V \to W$. It remains to 
  show that $V = \mathcal H(\R^d)$. $W$ is complete, so $V$ is.  Further, 
  $\mathcal{E}(\R^d) \subset V$, and by Fourier inversion each $f \in V$ 
  satisfies the reproducing property, \[f(x) = \int e^{2\pi i\langle x, 
  \xi\rangle} \widehat{f}(\xi) \, d\xi = \int \frac{\widehat{f}(\xi) 
  \overline{\widehat{k}_x(\xi)}}{\widehat{K}(\xi)} \, d\xi = \langle f, k_x 
  \rangle,\]
  so $\mathcal{H}(\R^d)$ is a closed subspace of $V$. Given $f \in 
  \mathcal{H}(\R^d)^\perp$, $f(x) = \langle f, k_x \rangle = 0$ for all $x$, 
  so $f = 0$. Thus $V = \mathcal{H}(\R^d)$.
\end{proof}

\begin{proof}[Proof of \autoref{lem:rkhs-embedding}]
  By \autoref{lem:rkhs-fourier}, the norm on $\mathcal H_\theta(\R^d)$ is
  \[\norm{f}^2_{\mathcal H_\theta(\R^d)} = \int 
  \frac{\abs{\widehat{f}(\xi)}^2}{\widehat{K}_\theta(\xi)} \, d\xi,\]
  and $K_\theta$ has Fourier transform
  \[\widehat K_\theta(\xi) = \frac{\widehat K(\xi_1/\theta_1, \dots, 
  \xi_d/\theta_d)}{\prod_{i=1}^d \theta_i}.\]
  If $\nu < \infty$, by assumption $\widehat{K}(\xi) = \widehat{k}(\norm{\xi})$, 
  for a finite non-increasing function $\widehat{k}$ satisfying 
  $\widehat{k}(\norm{\xi}) = \Theta(\norm{\xi}^{-2\nu-d})$ as $\xi \to \infty$.  
  Hence
  \[C(1 + \norm{\xi}^2)^{-(\nu+d/2)} \le \widehat{K}_\theta(\xi) \le C'(1 + 
  \norm{\xi}^2)^{-(\nu+d/2)},\]
  for constants $C,C' > 0$, and we obtain that $\mathcal{H}_\theta(\R^d)$ is 
  equivalent to the Sobolev space $H^{\nu+d/2}(\R^d)$.
  
  From \autoref{lem:rkhs-restriction}, $\mathcal{H}_\theta(D)$ is given by the 
  restriction of functions in $\mathcal{H}_\theta(\R^d)$; as $D$ is Lipschitz, 
  the same is true of $H^{\nu+d/2}$. $\mathcal{H}_\theta(D)$ is thus 
  equivalent to $H^{\nu+d/2}(D)$.  Finally, functions in 
  $\mathcal{H}_\theta(\bar{D})$ are continuous, so uniquely identified by 
  their restriction to $D$, and
  \[\mathcal{H}_\theta(\bar{D}) \simeq \mathcal{H}_\theta(D) \simeq 
  H^{\nu+d/2}(D).\]
  If $\nu = \infty$, by a similar argument $\mathcal{H}_\theta(\bar{D})$ is 
  continuously embedded in all $H^s(D)$.
\end{proof}

From \autoref{lem:rkhs-fourier}, we can derive results on the behaviour of 
$\norm{f}_{\mathcal{H}_\theta(S)}$ as $\theta$ varies.  For small $\theta$, we 
obtain the following result.

\begin{lemma}
  \label{lem:bounded-norm}
  If $f \in \mathcal{H}_{\theta}(S)$, then $f \in \mathcal{H}_{\theta'}(S)$ 
  for all $0 < \theta' \le \theta$, and \[\norm{f}_{\mathcal{H}_{\theta'}(S)}^2 
  \le \left(\prod_{i=1}^d \theta_i/\theta'_i\right) 
  \norm{f}_{\mathcal{H}_{\theta}(S)}^2.\]
\end{lemma}

\begin{proof}
  Let $C = \prod_{i=1}^d (\theta'_i/\theta_i)$. As $\widehat{K}$ is isotropic 
  and radially non-increasing,
  \[\widehat{K}_{\theta'}(\xi) = C 
  \widehat{K}_{\theta}((\theta'_1/\theta_1)\xi_1 , \dots, (\theta'_d / 
  \theta_d)\xi_d)  \ge C \widehat{K}_{\theta}(\xi).\]  Given $f \in 
  \mathcal{H}_{\theta}(S)$, let $g \in \mathcal{H}_{\theta}(\R^d)$ be its 
  minimum norm extension, as in \autoref{lem:rkhs-restriction}. By 
  \autoref{lem:rkhs-fourier},
  \[\norm{f}_{\mathcal{H}_{\theta'}(S)}^2 \le 
  \norm{g}_{\mathcal{H}_{\theta'}(\R^d)}^2 = \int 
  \frac{\abs{\widehat{g}}^2}{\widehat{K}_{\theta'}} \le \int 
  \frac{\abs{\widehat{g}}^2}{C\widehat{K}_{\theta}} =
  C^{-1} \norm{f}_{\mathcal{H}_{\theta}(S)}^2. \qedhere\]
\end{proof}

Likewise, for large $\theta$, we obtain the following.

\begin{lemma}
  \label{lem:norm-rates}
  If $\nu < \infty$, $f \in \mathcal{H}_{\theta}(S)$, then $f \in 
  \mathcal{H}_{t\theta}(S)$ for $t \ge 1$, and
  \[\norm{f}_{\mathcal{H}_{t\theta}(S)}^2 \le C'' 
  t^{2\nu}\norm{f}^2_{\mathcal{H}_\theta(S)},\]
  for a $C'' > 0$ depending only on $K$ and $\theta$.
\end{lemma}

\begin{proof}
  As in the proof of \autoref{lem:rkhs-embedding}, we have constants $C, C' > 
  0$ such that \[C(1 + \norm{\xi}^2)^{-(\nu + d/2)} \le 
  \widehat{K}_\theta(\xi) \le C'(1 + \norm{\xi}^2)^{-(\nu + d/2)}.\]
  Thus for $t \ge 1$,
  \begin{align*}
    \widehat{K}_{t\theta}(\xi) = t^{d}\widehat{K}_\theta(t\xi) 
    &\ge Ct^{d}(1+t^2\norm{\xi}^2)^{-(\nu+d/2)} \\
    &\ge Ct^{-2\nu}(1 + \norm{\xi}^2)^{-(\nu+d/2)} \\
    &\ge CC'^{-1}t^{-2\nu}\widehat{K}_\theta(\xi),
  \end{align*}
  and we may argue as in the previous lemma.
\end{proof}

We can also describe the posterior distribution of $f$ in terms of 
$\mathcal{H}_\theta(S)$; as a consequence, we may deduce \autoref{cor:r-hat}.

\begin{lemma}
  \label{lem:rkhs-posterior}
  Suppose $f(x) = \mu + g(x)$, $g \in \mathcal{H}_\theta(S)$.
  \begin{enumerate}
    \item $\hat{f}_n(x; \theta) = \hat{\mu}_n + \hat{g}_n(x)$ solves the 
      optimization problem
      \[\text{minimize } \norm{\hat{g}}_{\mathcal{H}_\theta(S)}^2, \quad 
      \text{subject to } \hat{\mu} + \hat{g}(x_i) = z_i, \quad 1 \le i \le 
      n,\]
      with minimum value $\hat{R}_n^2(\theta)$.
    \item The prediction error satisfies
      \[\abs{f(x) - \hat{f}_n(x; \theta)} \le 
      s_n(x;\theta)\norm{g}_{\mathcal{H}_\theta(S)}\]
      with equality for some $g \in \mathcal{H}_\theta(S)$.
  \end{enumerate}
\end{lemma}

\begin{proof}\ 

  \begin{enumerate}
    \item
      Let $W = \spn(k_{x_1}, \dots, k_{x_n})$, and write $\hat{g} = 
      \hat{g}^{\parallel} + \hat{g}^\perp$ for $\hat{g}^{\parallel} \in W$, 
      $\hat{g}^\perp \in W^\perp$.  $\hat{g}^\perp(x_i) = \langle 
      \hat{g}^\perp, k_{x_i} \rangle = 0$, so $\hat{g}^\perp$ affects the 
      optimization only through $\norm{\hat{g}}$. The minimal $\hat{g}$ thus 
      has $\hat{g}^\perp = 0$, so $\hat{g} = \sum_{i=1}^n \lambda_i k_{x_i}$.  
      The problem then becomes
      \[\text{minimize } \lambda^T V \lambda, \quad \text{subject to } 
      \hat{\mu}1 + V \lambda = z.\]
      The solution is given by \eqref{eq:pred-mu} and \eqref{eq:pred-mean}, 
      with value \eqref{eq:rss}.
    \item
      By symmetry, the prediction error does not depend on $\mu$, so we may 
      take $\mu = 0$. Then
      \[f(x) - \hat{f}_n(x; \theta) = g(x) - (\hat{\mu}_n + \hat{g}_n(x)) = 
      \langle g, e_{n,x} \rangle,\]
      for $e_{n,x} = k_x - \sum_{i=1}^n \lambda_i k_{x_i}$, and
      \[\lambda = \frac{V^{-1}1}{1^TV^{-1}1} + \left(I - 
      \frac{V^{-1}1}{1^TV^{-1}1}1^T\right)V^{-1}v.\]
      Now, $\norm{e_{n,x}}^2_{\mathcal{H}_\theta(S)} = s_n^2(x;\theta)$, as given 
      by \eqref{eq:pred-var}; this is a consequence of Lo\`{e}ve's isometry, 
      but is easily verified algebraically.  The result then follows by 
      Cauchy-Schwarz.  \qedhere
  \end{enumerate}
\end{proof}

\subsection{Fixed Parameters}

\begin{proof}[Proof of \autoref{thm:minimax-rates}]
  We first establish the lower bound. Suppose we have $2n$ functions $\psi_m$ 
  with disjoint supports. We will argue that, given $n$ observations, we 
  cannot distinguish between all the $\psi_m$, and thus cannot accurately pick 
  a minimum $x_n^*$.
  
  To begin with, assume $X = [0,1]^d$.  Let $\psi : \R^d \to [0,1]$ be a 
  $C^\infty$ function, supported inside $X$ and with minimum -1. By 
  \autoref{lem:rkhs-embedding}, $\psi \in \mathcal{H}_\theta(\R^d)$.  Fix $k 
  \in \N$, and set $n = (2k)^d/2$. For vectors $m \in \{0, \dots, 2k-1\}^d$, 
  construct functions $\psi_m(x) = C(2k)^{-\nu}\psi(2kx - m)$, where $C > 0$ 
  is to be determined.  $\psi_m$ is given by a translation and scaling of 
  $\psi$, so by {Lemmas \ref{lem:rkhs-fourier}}, 
  \ref{lem:rkhs-restriction} and \ref{lem:norm-rates}, for some $C' > 0$,
  \[\norm{\psi_m}_{\mathcal{H}_\theta(X)} \le 
  \norm{\psi_m}_{\mathcal{H}_\theta(\R^d)} = 
  C(2k)^{-\nu}\norm{\psi}_{\mathcal{H}_{2k\theta}(\R^d)} \le 
  CC'\norm{\psi}_{\mathcal{H}_{\theta}(\R^d)}.\]
  Set $C = R/C'\norm{\psi}_{\mathcal{H}_{\theta}(\R^d)}$, so that
  $\norm{\psi_m}_{\mathcal{H}_\theta(X)} \le R$ for all $m$ and $k$.

  Suppose $f = 0$, and let $x_n$ and $x_n^*$ be chosen by any valid strategy 
  $u$. Set $\chi = \{x_1, \dots, x_{n-1}, x_{n-1}^*\}$, and let $A_m$ be the 
  event that $\psi_m(x) = 0$ for all $x \in \chi$.  There are $n$ points in 
  $\chi$, and the $2n$ functions $\psi_m$ have disjoint support, so $\sum_m 
  \mathbb{I}(A_m) \ge n$. Thus
  \[\sum_m \P_0^u(A_m) = \E_0^u\left[\sum_m \mathbb{I}(A_m)\right] \ge n,\]
  and we have some fixed $m$, depending only on $u$, for which $\P_0^u(A_m) 
  \ge \frac12$.  On the event $A_m$,
  \[\psi_m(x_{n-1}^*) - \min \psi_m = C(2k)^{-\nu},\]
  but on that event, $u$ cannot distinguish between $0$ and $\psi_m$ before 
  time $n$, so
  \[C^{-1}(2k)^\nu\E_{\psi_m}^u[f(x_{n-1}^*) - \min f] \ge \P_{\psi_m}^u(A_m) 
  = \P_0^u(A_m) \ge \tfrac12.\]
  
  As the minimax loss is non-increasing in $n$, for $(2(k-1))^d/2 \le n < 
  (2k)^d/2$ we conclude
  \begin{align*}
    \inf_u L_n(u, \mathcal{H}_\theta(X), R) &\ge \inf_u L_{(2k)^d/2-1}(u, 
    \mathcal{H}_\theta(X), R) \\
    &\ge \inf_u \sup_m \E_{\psi_m}^u\left[f\left(x_{(2k)^d/2-1}^*\right)-\min 
    f\right]\\
    &\ge \tfrac12C(2k)^{-\nu} = \Omega(n^{-\nu/d}).
  \end{align*}
  For general $X$ having non-empty interior, we can find a hypercube $S = x_0 
  + [0, \varepsilon]^d \subseteq X$, with $\varepsilon > 0$.  We may then 
  proceed as above, picking functions $\psi_m$ supported inside $S$.

  For the upper bound, consider a strategy $u$ choosing a fixed sequence 
  $x_n$, independent of the $z_n$. Fit a radial basis function interpolant 
  $\hat{f}_n$ to the data, and pick $x_n^*$ to minimize $\hat f_n$. Then if 
  $x^*$ minimizes $f$,
  \begin{align*}
    f(x_n^*) - f(x^*) &\le f(x_n^*) - \hat f_n(x_n^*) + \hat f_n(x^*) - f(x^*)\\
    &\le 2\norm{\hat f_n - f}_\infty,
  \end{align*}
  so the loss is bounded by the error in $\hat f_n$.
  
  From results in \citet[\S6]{narcowich_refined_2003} and 
  \citet[\S11.5]{wendland_scattered_2005}, for suitable radial basis functions 
  the error is uniformly bounded by
  \[\sup_{\norm{f}_{\mathcal H_\theta(X)} \le R} \norm{\hat f_n - f}_\infty = 
  O(h_n^{-\nu}),\]
  where the mesh norm \[h_n \coloneqq \sup_{x \in X} \min_{i=1}^n \norm{x - 
  x_i}.\]
  (For $\nu \not\in \N$, this result is given by 
  \citeauthor{narcowich_refined_2003}\ for the radial basis function $K^\nu$, 
  which is $\nu$-H\"{o}lder at 0 by \citealp[\S9.6]{abramowitz_handbook_1965}; 
  for $\nu \in \N$, the result is given by 
  \citeauthor{wendland_scattered_2005} for thin-plate splines.)
  As $X$ is bounded, we may choose the $x_n$ so that $h_n = O(n^{-1/d})$, 
  giving
  \[L_n(u, H_\theta(X), R) = O(n^{-\nu/d}). \qedhere\]
\end{proof}

To prove \autoref{thm:ei-rates}, we first show that some observations $z_n$ 
will be well-predicted by past data.

\begin{lemma}
  \label{lem:var-upper-bound}
  
  Set
  \[\beta \coloneqq \begin{cases} \alpha, & \nu \le 1, \\ 0, & \nu > 1.  
  \end{cases}\]
  Given $\theta \in \R^d_+$, there is a constant $C' > 0$ depending only on 
  $X$, $K$ and $\theta$ which satisfies the following.  For any $k \in \N$, 
  and sequences $x_n \in X$, $\theta_n \ge \theta$, the inequality
  \[s_n(x_{n+1}; \theta_n) \ge C'k^{-(\nu \wedge 1)/d}(\log k)^{\beta}\]
  holds for at most $k$ distinct $n$.
\end{lemma}

\begin{proof}
  We first show that the posterior variance $s_n^2$ is bounded by the distance 
  to the nearest design point.  Let $\pi_n$ denote the prior with variance 
  $\sigma^2 = 1$, and length-scales $\theta_n$. Then for any $i \le n$, as 
  $\hat{f}_n(x; \theta_n) = \E_{\pi_n}[f(x) \mid \mathcal F_n]$,
  \begin{align*}
    s^2_n(x;\theta_n) &= \E_{\pi_n}[(f(x) - \hat f_n(x; \theta_n))^2 \mid 
    \mathcal{F}_n] \\
    &= \E_{\pi_n}[(f(x) - f(x_i))^2 - (f(x_i) - \hat f_n(x; \theta_n))^2 \mid 
    \mathcal{F}_n ]\\
    &\le \E_{\pi_n}[(f(x) - f(x_i))^2 \mid \mathcal{F}_n]\\
    &= 2(1 - K_{\theta_n}(x-x_i)).
  \end{align*}

  If $\nu \le \frac12$, then by assumption
  \[\abs{K(x) - K(0)} = O\left(\norm{x}^{2\nu} (-\log \norm{x})^{2\alpha}\right)\]
  as $x \to 0$.  If $\nu > \frac12$, then $K$ is differentiable, so as $K$ is 
  symmetric, $\nabla K(0) = 0$. If further $\nu \le 1$, then
  \[\abs{K(x)-K(0)} = \abs{K(x) - K(0) - x \cdot \nabla K(0)} = 
  O\left(\norm{x}^{2\nu} (-\log \norm{x})^{2\alpha}\right).\]
  Similarly, if $\nu > 1$, then $K$ is $C^2$, so
  \[\abs{K(x) - K(0)} = \abs{K(x) - K(0) - x \cdot \nabla K(0)} = 
  O(\norm{x}^{2}).\]
  We may thus conclude
  \[\abs{1 - K(x)} = \abs{K(x) - K(0)} = O\left(\norm{x}^{2(\nu \wedge 1)} 
  (-\log \norm{x})^{2\beta}\right),\]
  and
  \[s^2_n(x; \theta_n) \le C^2\norm{x-x_i}^{2(\nu \wedge 1)} (-\log \norm{x - 
  x_i})^{2\beta},\]
  for a constant $C > 0$ depending only on $X$, $K$ and $\theta$.
  
  We next show that most design points $x_{n+1}$ are close to a previous 
  $x_i$.  $X$ is bounded, so can be covered by $k$ balls of radius 
  $O(k^{-1/d})$.  If $x_{n+1}$ lies in a ball containing some earlier point 
  $x_i$, $i \le n$, then we may conclude
  \[s^2_n(x_{n+1}; \theta_n) \le C'^2k^{-2(\nu \wedge 1)/d} (\log 
  k)^{2\beta},\]
  for a constant $C' > 0$ depending only on $X$, $K$ and $\theta$. Hence as 
  there are $k$ balls, at most $k$ points $x_{n+1}$ can satisfy
  \[s_n(x_{n+1};\theta_n) \ge C'k^{-(\nu \wedge 1)/d} (\log k)^{\beta}.  
  \qedhere\]
\end{proof}

Next, we provide bounds on the expected improvement when $f$ lies in the RKHS.

\begin{lemma}
  \label{lem:ei-rates}

  Let $\norm{f}_{\mathcal{H}_\theta(X)} \le R$.  For $x \in X$, $n \in \N$, set 
  $I = (f(x_n^*) - f(x))^+$, and $s = s_n(x;\theta)$. Then for
  \[\tau(x) \coloneqq x\Phi(x) + \phi(x),\]
  we have
  \[\max\left(I - Rs, \frac{\tau(-R/\sigma)}{\tau(R/\sigma)}I\right) \le 
  EI_n(x; \pi) \le I + (R + \sigma)s.\]
\end{lemma}

\begin{proof}
  If $s = 0$, then by \autoref{lem:rkhs-posterior}, $\hat f_n(x; \theta) = 
  f(x)$, so $EI_n(x; \pi) = I$, and the result is trivial.  Suppose $s > 0$, 
  and set $t = (f(x_n^*) - f(x)) / s$, $u = (f(x_n^*) - \hat f_n(x; \theta)) / 
  s$.  From \eqref{eq:ei-formula} and \eqref{eq:ei-rho},
  \[EI_n(x; \pi) = \sigma s\tau(u/\sigma),\]
  and by \autoref{lem:rkhs-posterior}, $\abs{u - t} \le R$. As $\tau'(z) = \Phi(z) 
  \in [0,1]$, $\tau$ is non-decreasing, and $\tau(z) \le 1 + z$ for $z \ge 0$.  
  Hence
  \[
  EI_n(x;\pi) \le \sigma s \tau\left(\frac{t^++R}{\sigma}\right)
  \le \sigma s \left(\frac{t^++R}{\sigma}+1\right)
  = I + (R+ \sigma)s. \]

  If $I = 0$, then as $EI$ is the expectation of a non-negative quantity, $EI 
  \ge 0$, and the lower bounds are trivial. Suppose $I > 0$.  Then as $EI \ge 
  0$, $\tau(z) \ge 0$ for all $z$, and $\tau(z) = z + \tau(-z) \ge z$.  Thus 
  \[EI_n(x;\pi) \ge \sigma s \tau\left(\frac{t-R}\sigma\right) \ge \sigma s 
  \left(\frac{t-R}\sigma\right) = I - Rs.\]
  Also, as $\tau$ is increasing,
  \[EI_n(x;\pi) \ge \sigma \tau\left(\frac{-R}\sigma\right) s.\]
  Combining these bounds, and eliminating $s$, we obtain
  \[EI_n(x;\pi) \ge \frac{\sigma \tau(-R/\sigma)}{R + \sigma \tau(-R/\sigma)} 
  I = \frac{\tau(-R/\sigma)}{\tau(R/\sigma)}I. \qedhere\]
\end{proof}

We may now prove the theorem. We will use the above bounds to show that there 
must be times $n_k$ when the expected improvement is low, and thus 
$f(x_{n_k}^*)$ is close to $\min f$.

\begin{proof}[Proof of \autoref{thm:ei-rates}]
  From \autoref{lem:var-upper-bound} there exists $C > 0$, depending on $X$, 
  $K$ and $\theta$, such that for any sequence $x_n \in X$ and $k \in \N$, the 
  inequality
  \[s_n(x_{n+1}; \theta) > Ck^{-(\nu \wedge 1)/d}(\log k)^{\beta}\] holds at 
  most $k$ times.  Furthermore, $z_n^* - z_{n+1}^* \ge 0$, and for 
  $\norm{f}_{\mathcal{H}_\theta(X)} \le R$,
  \[\sum_n z_n^* - z_{n+1}^* \le z_1^* - \min f \le 2\norm{f}_\infty \le 2R,\]
  so $z_n^* - z_{n+1}^* > 2R k^{-1}$ at most $k$ times. Since $z_n^* - 
  f(x_{n+1}) \le z_n^* - z_{n+1}^*$, we have also $z_n^* - f(x_{n+1}) > 2R 
  k^{-1}$ at most $k$ times.  Thus there is a time $n_k$, $k \le n_k \le 3k$, 
  for which $s_{n_k}(x_{n_k+1}; \theta) \le Ck^{-(\nu \wedge 1)/d}(\log 
  k)^{\beta}$ and $z_{n_k}^* - f(x_{n_k+1}) \le 2R k^{-1}$.
 
  Let $f$ have minimum $z^*$ at $x^*$. For $k$ large, $x_{n_k+1}$ will have 
  been chosen by expected improvement (rather than being an initial design 
  point, chosen at random). Then as $z_n^*$ is non-increasing in $n$, for $3k 
  \le n < 3(k+1)$ we have by \autoref{lem:ei-rates},
  \begin{align*}
    z_n^* -z^* &\le z_{n_k}^* - z^*\\
    &\le \frac{\tau(R/\sigma)}{\tau(-R/\sigma)} EI_{n_k}(x^*; \pi)\\
    &\le \frac{\tau(R/\sigma)}{\tau(-R/\sigma)} EI_{n_k}(x_{n_k+1}; \pi)\\
    &\le \frac{\tau(R/\sigma)}{\tau(-R/\sigma)} \left( 2Rk^{-1} + C(R + 
    \sigma)k^{-(\nu \wedge 1)/d}(\log k)^{\beta}\right).
  \end{align*}
  This bound is uniform in $f$ with $\norm{f}_{\mathcal H_\theta(X)} \le R$, so 
  we obtain
  \[L_n(EI(\pi), \mathcal{H}_\theta(X), R)  = O(n^{-(\nu \wedge 1)/d}(\log 
  n)^\beta). \qedhere\]
\end{proof}

\subsection{Estimated Parameters}

To prove \autoref{thm:ei-hat-diverges}, we first establish lower bounds on the 
posterior variance.

\begin{lemma}
  \label{lem:var-lower-bound}

  Given $\theta^L, \theta^U \in \R^d_+$, pick sequences $x_n \in X$, $\theta^L 
  \le \theta_n \le \theta^U$. Then for open $S \subset X$,
  \[\sup_{x \in S} s_n(x; \theta_n) = \Omega(n^{-\nu/d}),\]
  uniformly in the sequences $x_n$, $\theta_n$.
\end{lemma}

\begin{proof}
  $S$ is open, so contains a hypercube $T$. For $k \in \N$, let $n = 
  \frac12(2k)^d$, and construct $2n$ functions $\psi_m$ on $T$ with 
  $\norm{\psi_m}_{\mathcal{H}_{\theta^U}(X)} \le 1$, as in the proof of 
  \autoref{thm:minimax-rates}. Let $C^2 = \prod_{i=1}^d 
  (\theta^U_i/\theta^L_i)$; then by \autoref{lem:bounded-norm}, 
  $\norm{\psi_m}_{\mathcal{H}_{\theta_n}(X)} \le C$.
      
  Given $n$ design points $x_1, \dots, x_n$, there must be some $\psi_m$ such 
  that $\psi_m(x_i) = 0$, $1 \le i \le n$.  By \autoref{lem:rkhs-posterior}, 
  the posterior mean of $\psi_m$ given these observations is the zero 
  function.  Thus for $x \in T$ minimizing $\psi_m$,
  \[s_n(x;\theta_n) \ge 
  C^{-1}s_n(x;\theta_n)\norm{\psi_m}_{\mathcal{H}_{\theta_n}(X)}
  \ge C^{-1}\abs{\psi_m(x) - 0} = \Omega(k^{-\nu}).\]
  As $s_n(x; \theta)$ is non-increasing in $n$, for $\frac12(2(k-1))^d < n 
  \le \frac12(2k)^d$ we obtain
  \[\sup_{x \in S} s_n(x; \theta_n) \ge \sup_{x \in S} s_{\frac12(2k)^d} (x; 
  \theta_n) = \Omega(k^{-\nu}) = \Omega(n^{-\nu/d}). \qedhere\]
\end{proof}

Next, we bound the expected improvement when prior parameters are estimated by 
maximum likelihood.

\begin{lemma}
  \label{lem:ei-hat-rates}
  Let $\norm{f}_{\mathcal{H}_{\theta^U}(X)} \le R$, $x_n, y_n \in X$.  Set 
  $I_n(x) = z_n^* - f(x)$, $s_n(x) = s_n(x; \hat{\theta}_n)$, and $t_n(x) = 
  I_n(x)/s_n(x)$. Suppose:
  \begin{enumerate}
    \item for some $i < j$, $z_i \ne z_j$;
    \item for some $T_n \to -\infty$, $t_n(x_{n+1}) \le T_n$ whenever 
      $s_n(x_{n+1}) > 0$;
    \item $I_n(y_{n+1}) \ge 0$; and
    \item for some $C > 0$, $s_n(y_{n+1}) \ge e^{-C/c_n}$.
  \end{enumerate}
  Then for $\hat{\pi}_n$ as in \autoref{def:ei-hat}, eventually $EI_n(x_{n+1}; 
  \hat{\pi}_n) < EI_n(y_{n+1}; \hat{\pi}_n)$.
  If the conditions hold on a subsequence, so does the conclusion.
\end{lemma}

\begin{proof}
  Let $\hat{R}_n^2(\theta)$ be given by \eqref{eq:rss}, and set $\hat{R}_n^2 = 
  \hat{R}_n^2(\hat{\theta}_n)$. For $n \ge j$, $\hat{R}_n^2 > 0$, and by 
  \autoref{lem:bounded-norm} and \autoref{cor:r-hat},
  \[\hat{R}_n^2 \le \norm{f}^2_{\mathcal{H}_{\hat{\theta}_n}(X)} \le S^2 = R^2 
  \prod_{i=1}^d (\theta^U_i/\theta^L_i).\]
  Thus $0 < \hat{\sigma}^2_n \le S^2c_n$.  Then if $s_n(x) > 0$, for some 
  $\abs{u_n(x) - t_n(x)} \le S$,
  \[EI_n(x; \hat{\pi}_n) = \hat{\sigma}_n s_n(x) 
  \tau(u_n(x)/\hat{\sigma}_n),\]
  as in the proof of \autoref{lem:ei-rates}.

  If $s_n(x_{n+1}) = 0$, then $x_{n+1} \in \{x_1, \dots, x_n\}$, so 
  \[EI_n(x_{n+1}; \hat{\pi}_n) = 0 < EI_n(y_{n+1}; \hat{\pi}_n).\]
  When $s_n(x_{n+1}) > 0$, as $\tau$ is increasing we may upper bound 
  $EI_n(x_{n+1}; \hat{\pi}_n)$ using $u_n(x_{n+1}) \le T_n + S$, and lower 
  bound $EI_n(y_{n+1}; \hat{\pi}_n)$ using $u_n(y_{n+1}) \ge -S$. Since 
  $s_n(x_{n+1}) \le 1$, and
  $\tau(x) = \Theta(x^{-2}e^{-x^2/2})$ as $x \to -\infty$ 
  \citep[\S7.1]{abramowitz_handbook_1965},
  \begin{align*}
    \frac{EI_n(x_{n+1}; \hat{\pi}_n)}{EI_n(y_{n+1}; \hat{\pi}_n)} &\le
    \frac{\tau( (T_n+S)/\hat{\sigma}_n)}{e^{-C/c_n}\tau(-S/\hat{\sigma}_n)}\\
    &= O\left((T_n+S)^{-2} e^{C/c_n -( T_n^2 + 
    2ST_n)/2\hat{\sigma}_n^2}\right) \\
    &= O\left((T_n+S)^{-2} e^{-(T_n^2 + 2ST_n-2CS^2)/2S^2c_n} \right) \\
    &= o(1).
  \end{align*}
  If the conditions hold on a subsequence, we may similarly argue along that 
  subsequence.
\end{proof}

Finally, we will require the following technical lemma.

\begin{lemma}
  \label{lem:points-open-set}
  Let $x_1, \dots, x_n$ be random variables taking values in $\R^d$. Given 
  open $S \subseteq \R^d$, there exist open $U \subseteq S$ for which 
  $\P(\bigcup_{i=1}^n\{x_i \in U\})$ is arbitrarily small.
\end{lemma}

\begin{proof}
  Given $\varepsilon > 0$, fix $m \ge n/\varepsilon$, and pick disjoint open 
  sets $U_1,\dots,U_m \subset S$. Then
  \[\sum_{j=1}^m \E[\#\{x_i \in U_j\}] \le \E[\#\{x_i \in \R^d\}] = n,\]
  so there exists $U_j$ with
  \[\P\left(\bigcup_i\{x_i \in U_j\}\right) \le \E[\#\{x_i \in U_j\}] \le n/m 
  \le \varepsilon. \qedhere\]
\end{proof}

We may now prove the theorem. We will construct a function $f$ on which the 
$EI(\hat \pi)$ strategy never observes within a region $W$. We may then 
construct a function $g$, agreeing with $f$ except on $W$, but having 
different minimum.  As the strategy cannot distinguish between $f$ and $g$, it 
cannot successfully find the minimum of both.

\begin{proof}[Proof of \autoref{thm:ei-hat-diverges}]
  Let the $EI(\hat{\pi})$ strategy choose initial design points $x_1, \dots, 
  x_k$, independently of $f$. Given $\varepsilon > 0$, by 
  \autoref{lem:points-open-set} there exists open $U_0 \subseteq X$ for which 
  $\P^{EI(\hat{\pi})}(x_1, \dots, x_k \in U_0) \le \varepsilon$; we may choose 
  $U_0$ so that  $V_0 = X \setminus U_0$ has non-empty interior.  Pick open 
  $U_1$ such that $V_1 = \bar{U}_1 \subset U_0$, and set $f$ to be a 
  $C^\infty$ function, 0 on $V_0$, 1 on $V_1$, and everywhere non-negative.  
  By \autoref{lem:rkhs-fourier}, $f \in \mathcal{H}_{\theta^U}(X)$.
  
  We work conditional on the event $A$, having probability at least 
  $1-\varepsilon$, that $z^*_k= 0$, and thus $z^*_n = 0$ for all $n \ge k$.  
  Suppose $x_n \in V_1$ infinitely often, so the $z_n$ are not all equal.  By 
  \autoref{lem:var-upper-bound}, $s_n(x_{n+1}; \hat{\theta}_n) \to 0$, so on a 
  subsequence with $x_{n+1} \in V_1$, we have
  \[t_n = (z_n^* - f(x_{n+1}))/s_n(x_{n+1}; \hat{\theta}_n) = -s_n(x_{n+1}; 
  \hat{\theta}_n)^{-1} \to -\infty\]
  whenever $s_n(x_{n+1}; \hat{\theta}_n) > 0$.
  However, by \autoref{lem:var-lower-bound}, there are points $y_n \in V_0$ 
  with $z_n^* - f(y_{n+1}) = 0$, and $s_n(y_{n+1}; \hat{\theta}_n) = 
  \Omega(n^{-\nu/d})$. Hence by \autoref{lem:ei-hat-rates}, $EI_n(x_{n+1}; 
  \hat{\pi}_n) < EI_n(y_{n+1}; \hat{\pi}_n)$ for some $n$, contradicting the 
  definition of $x_{n+1}$.

  Hence, on $A$, there is a random variable $T$ taking values in $\N$, for 
  which $n > T \implies x_n \not\in V_1$. Hence there exists a constant $t \in 
  \N$ for which the event $B = A \cap \{T \le t\}$ has 
  $\P_f^{EI(\hat{\pi})}$-probability at least $1 - 2\varepsilon$.  By 
  \autoref{lem:points-open-set}, we thus have an open set $W \subset V_1$ for 
  which the event \[C = B \cap \{x_n \not\in W : n \in \N\} = B \cap \{x_n 
  \not\in W : n \le t\}\] has $\P_f^{EI(\hat{\pi})}$-probability at least $1 - 
  3\varepsilon$.

  Construct a smooth function $g$ by adding to $f$ a $C^\infty$ function which 
  is 0 outside $W$, and has minimum $-2$. Then $\min g = -1$, but on the event 
  $C$, $EI(\hat{\pi})$ cannot distinguish between $f$ and $g$, and $g(x_n^*) 
  \ge 0$. Thus for $\delta = 1$,
  \[\P_g^{EI(\hat{\pi})}\left(\inf_n g(x_n^*) - \min g \ge \delta\right) \ge 
  \P_g^{EI(\hat{\pi})}(C) = \P_f^{EI(\hat{\pi})}(C) \ge 1-3\varepsilon.\]
  As the behaviour of $EI(\hat{\pi})$ is invariant under rescaling, we may 
  scale $g$ to have norm $\norm{g}_{\mathcal{H}_\theta(X)} \le R$, and the above 
  remains true for some $\delta > 0$.
\end{proof}

\begin{proof}[Proof of \autoref{thm:ei-tilde-rates}]
  As in the proof of \autoref{thm:ei-rates}, we will show there are times 
  $n_k$ when the expected improvement is small, so $f(x_{n_k})$ must be close 
  to the minimum. First, however, we must control the estimated parameters 
  $\hat \sigma^2_n$, $\hat \theta_n$.

  If the $z_n$ are all equal, then by assumption the $x_n$ are dense in $X$, 
  so $f$ is constant, and the result is trivial. Suppose the $z_n$ are not all 
  equal, and let $T$ be a random variable satisfying $z_T \ne z_i$ for some $i 
  < T$.  Set $U = \inf_{\theta^L \le \theta \le \theta^U} \hat{R}_T(\theta)$.  
  $\hat{R}_T(\theta)$ is a continuous positive function, so $U > 0$.  Let $S^2 
  = R^2\prod_{i=1}^d (\theta^U_i/\theta^L_i)$. By \autoref{lem:bounded-norm}, 
  $\norm{f}_{\mathcal{H}_{\hat{\theta}_n}(X)} \le S$, so by \autoref{cor:r-hat}, 
  for $n \ge T$, \[U \le \hat{R}_T(\hat{\theta}_n) \le \hat{\sigma}_n \le 
  \norm{f}_{\mathcal{H}_{\hat{\theta}_n}(X)} \le S.\]

  As in the proof of \autoref{thm:ei-rates}, we have a constant $C > 0$, and 
  some $n_k$, $k \le n_k \le 3k$, for which $z_{n_k}^* - f(x_{n_k+1}) \le 
  2Rk^{-1}$ and $s_{n_k}(x_{n_k+1}; \hat{\theta}_{n_k}) \le Ck^{-\alpha}(\log 
  k)^{\beta}$.  Then for $k \ge T$, $3k \le n < 3(k+1)$, arguing as in 
  \autoref{thm:ei-rates} we obtain
  \begin{align*}
  z_n^* - z^* &\le z_{n_k}^* - z^* \\
  &\le \frac{\tau(S/\hat{\sigma}_{n_k})}{\tau(-S/\hat{\sigma}_{n_k})} \left( 
  2Rk^{-1} + C(S + \hat{\sigma}_{n_k})k^{-(\nu \wedge 1)/d}(\log 
  k)^{\beta}\right)\\
  &\le \frac{\tau(S/U)}{\tau(-S/U)} \left( 2Rk^{-1} + 2CSk^{-(\nu \wedge 
  1)/d}(\log k)^{\beta}\right).
  \end{align*}
  We thus have a random variable $C'$ satisfying $z_n^* - z^* \le C'n^{-(\nu 
  \wedge 1)/d}(\log n)^\beta$ for all $n$, and the result follows.
\end{proof}

\subsection{Near-Optimal Rates}

To prove \autoref{thm:ei-epsilon-optimal}, we first show that the points chosen 
at random will be quasi-uniform in $X$.

\begin{lemma}
  \label{lem:random-search-uniform}

  Let $x_n$ be i.i.d.\ random variables, distributed uniformly over $X$, and 
  define their mesh norm,
  \[h_n \coloneqq \sup_{x \in X} \min_{i=1}^n \norm{x - x_i}.\]
  For any $\gamma > 0$, there exists $C > 0$ such that
  \[\P(h_n > C(n/\log n)^{-1/d}) = O(n^{-\gamma}).\]
\end{lemma}

\begin{proof}
  We will partition $X$ into $n$ regions of size $O(n^{-1/d})$, and show that 
  with high probability we will place an $x_i$ in each one. Then every point 
  $x$ will be close to an $x_i$, and the mesh norm will be small. 

  Suppose $X = [0, 1]^d$, fix $k \in \N$, and divide $X$ into $n = k^d$ 
  sub-cubes $X_m = \frac1k(m + [0, 1]^d)$, for $m \in \{0, \dots, k-1\}^d$.  
  Let $I_m$ be the indicator function of the event \[\{x_i \not \in X_m : 1 
  \le i \le \lfloor \gamma n \log n \rfloor\},\] and define
  \[\mu_n = \E \left[ \sum_m I_m \right] = n \E [I_0] = n(1-1/n)^{\lfloor 
  \gamma n \log n \rfloor} \sim ne^{-\gamma\log n} = n^{-(\gamma-1)}.\] For 
  $n$ large, $\mu_n \le 1$, so by the generalized Chernoff bound of 
  \citet[\S3.1]{panconesi_randomized_1997},
  \[
    \P\left(\sum_m  I_m \ge 1\right)
    \le \left(\frac{e^{(\mu_n^{-1}-1)}}{\mu_n^{-\mu_n^{-1}}}\right)^{\mu_n}
    \le e\mu_n \sim en^{-(\gamma-1)}.
    \]

  On the event $\sum_m I_m < 1$, $I_m = 0$ for all $m$. For any $x \in X$, we 
  then have $x \in X_m$ for some $m$, and $x_j \in X_m$ for some $1 \le j \le 
  \lfloor \gamma n \log n \rfloor$.  Thus
  \[\min_{i=1}^{\lfloor \gamma n \log n \rfloor} \norm{x - x_i} \le \norm{x - 
  x_j} \le \sqrt{d}k^{-1}.\]
  As this bound is uniform in $x$, we obtain $h_{\lfloor \gamma n \log n 
  \rfloor} \le \sqrt{d}k^{-1}$. Thus for $n = k^d$,
  \[\P(h_{\lfloor \gamma n \log n \rfloor} > \sqrt{d}k^{-1}) = O(k^{-d(\gamma 
  - 1)}),\]
  and as $h_n$ is non-increasing in $n$, this bound holds also for $k^d \le n 
  < (k+1)^d$.  By a change of variables, we then obtain
  \[\P(h_n > C(n / \gamma \log n)^{-1/d}) = O((n / \gamma \log 
  n)^{-(\gamma-1)}),\]
  and the result follows by choosing $\gamma$ large.  For general $X$, as $X$ 
  is bounded it can be partitioned into $n$ regions of measure 
  $\Theta(n^{-1/d})$, so we may argue similarly.
\end{proof}

We may now prove the theorem. We will show that the points $x_n$ must be 
quasi-uniform in $X$, so posterior variances must be small. Then, as in the 
proofs of
Theorems~\ref{thm:ei-rates} and 
\ref{thm:ei-tilde-rates}, we have times when the expected improvement is 
small, so $f(x_n^*)$ is close to $\min f$.

\begin{proof}[Proof of \autoref{thm:ei-epsilon-optimal}]
  First suppose $\nu < \infty$.  Let the $EI(\,\cdot\,, \varepsilon)$ choose 
  $k$ initial design points independent of $f$, and suppose $n \ge 2k$. Let 
  $A_n$ be the event that $\lfloor \frac{\varepsilon}{4}n \rfloor$ of the 
  points $x_{k+1}, \dots, x_n$ are chosen uniformly at random, so by a 
  Chernoff bound, \[\P^{EI(\,\cdot\,, \varepsilon)}(A_n^c) \le e^{-\varepsilon 
  n/16}.\] Let $B_n$ be the event that one of the points $x_{n+1}, \dots, 
  x_{2n}$ is chosen by expected improvement, so \[\P^{EI(\,\cdot\,, 
  \varepsilon)}(B_n^c) = \varepsilon^n.\] Finally, let $C_n$ be the event that 
  $A_n$ and $B_n$ occur, and further the mesh norm $h_n \le C(n/\log 
  n)^{-1/d}$, for the constant $C$ from \autoref{lem:random-search-uniform}.  
  Set $r_n = (n/\log n)^{-\nu/d}(\log n)^\alpha$.  Then by 
  \autoref{lem:random-search-uniform}, since $C_n \subset A_n$, 
  \[\P_f^{EI(\,\cdot\,, \varepsilon)}(C_n^c) \le C'r_n,\] for a constant $C' > 
  0$ not depending on $f$.

  Let $EI(\,\cdot\,, \varepsilon)$ have prior $\pi_n$ at time $n$, with (fixed 
  or estimated) parameters $\sigma_n$, $\theta_n$.  Suppose 
  $\norm{f}_{\mathcal{H}_{\theta^U}(X)} \le R$, and set $S^2 = R^2 
  \prod_{i=1}^d (\theta^U_i/\theta^L_i)$, so by \autoref{lem:bounded-norm}, 
  $\norm{f}_{\mathcal{H}_{\theta_n}(X)} \le S$.  If $\alpha = 0$, then by 
  \citet[\S6]{narcowich_refined_2003},
  \[\sup_{x \in X} s_n(x; \theta) = O(M(\theta)h^\nu_n)\]
  uniformly in $\theta$, for $M(\theta)$ a continuous function of $\theta$.  
  Hence on the event $C_n$,
  \[\sup_{x \in X} s_n(x; \theta_n) \le \sup_{x \in X} \sup_{\theta^L \le 
  \theta \le \theta^U} s_n(x; \theta) \le C''r_n,\]
  for a constant $C'' > 0$ depending only on $X$, $K$, $C$, $\theta^L$ and 
  $\theta^U$.  If $\alpha > 0$, the same result holds by a similar argument.

  On the event $C_n$, we have some $x_m$ chosen by expected improvement, $n < 
  m \le 2n$. Let $f$ have minimum $z^*$ at $x^*$.  Then by 
  \autoref{lem:ei-rates},
  \begin{align*}
    z_{m-1}^* - z^* &\le EI_{m-1}(x^*; \,\cdot\,) + C''Sr_{m-1}\\
    &\le EI_{m-1}(x_m; \,\cdot\,) + C''Sr_{m-1} \\
    &\le (f(x_{m-1}) - f(x_m))^+ + C''(2S+\sigma_{m-1})r_{m-1}\\
    &\le z_{m-1}^* - z_m^* + C''Tr_n,
  \end{align*}
  for a constant $T > 0$. (Under $EI(\pi, \varepsilon)$, we have $T = 2S + 
  \sigma$; otherwise $\sigma_{m-1} \le S$ by \autoref{cor:r-hat}, so $T = 
  3S$.)
  Thus, rearranging,
  \[z_{2n}^* - z^* \le z_m^* - z^* \le C''Tr_n.\]

  On the event $C_n^c$, we have $z_{2n}^* - z^* \le 2\norm{f}_\infty \le 2R$, so
  \begin{align*}
    \E_f^{EI(\,\cdot\,, \varepsilon)}[z_{2n+1}^* -z^*] &\le 
    \E_f^{EI(\,\cdot\,, \varepsilon)}[z_{2n}^*-z^*] \\ &\le 
    2R\P_f^{EI(\,\cdot\,, \varepsilon)}(C_n^c) + C''Tr_n \\ &\le (2C'R + 
    C''T)r_n.
  \end{align*}
  As this bound is uniform in $f$ with $\norm{f}_{\mathcal{H}_{\theta^U}(X)} \le 
  R$, the result follows. If instead $\nu = \infty$, the above argument holds 
  for any $\nu < \infty$.
\end{proof}

\bibliographystyle{abbrvnat}
{\footnotesize \bibliography{cregoa}}

\end{document}